\documentclass[review]{elsarticle}

\usepackage{lineno,hyperref}
\graphicspath{{figures/}{../figures/}}
\usepackage{subfiles}
\usepackage{multirow}
\usepackage{caption}
\usepackage{subcaption}
\usepackage{xcolor}
\usepackage{soul}
\usepackage{float}
\modulolinenumbers[5]

\journal{Expert Systems with Applications}

\biboptions{authoryear}

\begin{document}

\begin{frontmatter}

\title{Player Identification in Hockey Broadcast Videos\tnoteref{mytitlenote}}
\tnotetext[mytitlenote]{This work was supported in part by the Natural Sciences and Engineering Research Council of Canada (NSERC), Sportlogiq and the Mitacs Accelerate program. The GPUs, data, and other resources used in this research project were provided by Sportlogiq. All material related to Sportlogiq intellectual property is presented in this paper with permission.}

\author[address1]{Alvin Chan\corref{mycorrespondingauthor}}
\cortext[mycorrespondingauthor]{Corresponding author}
\ead{alvin.chan3@mail.mcgill.ca}

\author[address1]{Martin D. Levine}
\author[address1,address2]{Mehrsan Javan}

\address[address1]{McGill University, 845 Rue Sherbrooke O,  Montreal, Quebec, H3A 0G4, Canada}
\address[address2]{Sportlogiq, 5455 Avenue de Gasp\'e Suite 570, Montreal, Quebec, H2T 3B3, Canada}

\begin{abstract}
We present a deep recurrent convolutional neural network (CNN) approach to solve the problem of hockey player identification in NHL broadcast videos. Player identification is a difficult computer vision problem mainly because of the players’ similar appearance, occlusion, and blurry facial and physical features. However, we can observe players' jersey numbers over time by processing variable length image sequences of players (aka ‘tracklets’). We propose an end-to-end trainable ResNet+LSTM network, with a residual network (ResNet) base and a long short-term memory (LSTM) layer, to discover spatio-temporal features of jersey numbers over time and learn long-term dependencies. Additionally, we employ a secondary 1-dimensional convolutional neural network classifier as a late score-level fusion method to classify the output of the ResNet+LSTM network. For this work, we created a new hockey player tracklet dataset that contains sequences of hockey player bounding boxes. This achieves an overall player identification accuracy score over 87\% on the test split of our new dataset. 
\end{abstract}


\end{frontmatter}


\section{Introduction}

Ice hockey is a widely popular sport in North America that attracts millions of television viewers to National Hockey League (NHL) games. Recently, sports analytics have taken the sports industry by storm and have become a major focus for the NHL. Sports analytics provides greater insights into the players' abilities and the game plays, which are indispensable to coaches for making better informed decisions to fully utilize their players' potential. This information can also be incorporated into TV broadcast videos to enhance a viewer’s experience. More interesting storylines can be told and supported with actual data rather than relying on a coach's or analyst's opinion. However, hockey games are currently annotated manually to track and identify players and to record all events that occur throughout the game. This is laborious and time-intensive. A fully automatic player detection, tracking, and identification system would facilitate the process and is highly sought after in the sports industry.

In this work, we aim to contribute to building a full sport analytics system from the TV broadcast feed by focusing on the task of automatic hockey player identification in NHL broadcast videos. \textit{NHL broadcast videos are primarily recorded from the main camera, which offers the standard view of the rink.} The main challenge is that teammates have very similar appearances and their facial and physical body features in the broadcast video are almost indistinguishable from each other, due to the limited frame resolution of the camera. As players skate around the rink, they also become blurry and often occlude each other from the camera's view. The most unique observable feature of each player would be the jersey number on their back, though it is not always visible due to occlusion, motion blur, and the angle in which the players are facing the camera. Our solution to this problem is to take temporal information across multiple frames and identify a player by predicting the jersey number. For the purpose of this research, we only consider sections of the NHL broadcast videos that are recorded by the main camera.

As a project supported by Sportlogiq~\footnote{http://sportlogiq.com/en/}, the player identification system we develop is an extension to their existing player detection and tracking system. This system generates player tracking data from the broadcast feed to provide player location information, which we can use to extract bounding boxes around players seen in each video frame, thereby creating image sequences of players. Given an image sequence of a player, our objective is to identify the jersey number of the player within the image sequence and assign the sequence with correct jersey number label. Note that our current iteration of the player identification system runs offline, but is applicable to any player image sequence regardless of how the bounding boxes are extracted.

Encouraged by the state-of-the-art results in street number recognition ~\citep{Goodfellow2013}, we apply convolutional neural networks (CNN) to perform jersey number recognition. Since we simultaneously consider multiple images, we employ a recurrent CNN, similar to the "long-term recurrent convolutional network" (LRCN) ~\citep{Donahue2016}. A recurrent CNN can process image sequences by first using convolutional layers to extract a feature vector representation for each image. Subsequently, the "long short-term memory" (LSTM)~\citep{Hochreiter1997} units in the recurrent layer of the network permit information to flow across time-steps and enable the use of information from previous images. 

Our contribution to the literature is an approach that identifies hockey players from a single camera broadcast video feed by combining a CNN with a LSTM layer to discover spatio-temporal features in an image sequence of players (tracklet). Additionally, we demonstrate the use of a second CNN in a score-level fusion for the network outputs over a tracklet to generate a more accurate final prediction.

\section{Related Work}

\subsection{Background}
CNN have achieved significant progress in recent years and have become the backbone in almost all approaches to visual image recognition problems. Their popularity grew after they were implemented by~\cite{Krizhevsky2012} to obtain state-of-the-art results on the ImageNet Large-Scale Visual Recognition Challenge 2012 ~\citep{ILSVRC15}. Recurrent neural networks (RNN) were first introduced in the literature~\citep{Rumelhart1986, Werbos1988} in the 1980s and were developed as a method for modeling data points that occur in a time series. In theory, the ordinary RNN is an ideal model for processing sequential data, but it is difficult to train in practice for learning long-term dynamics. This is attributed to occurrences known as the "exploding gradient" and "vanishing gradient"~\citep{Bengio1994}. \cite{Pascanu2012} showed that the exploding gradient problem can be mitigated by clipping the gradient at a maximum threshold. However, in order to alleviate the vanishing gradient issue, ~\cite{Hochreiter1997} proposed the long short-term memory (LSTM) to ease the difficulties of training RNNs~\citep{Bengio1994}.

\subsection{Sports Player Identification}
At the time of this paper, there has only been a handful of works relevant to sports player identification. The majority of these have been developed for videos of soccer and basketball games. To the best of our knowledge, our proposed method is the first to solve the problem of player identification in hockey broadcast videos.

Most of the earlier computer vision approaches rely on close-up shots of players in a broadcast video feed. The majority of these follow similar methods using optical character recognition (OCR) software to distinguish players by their jersey number, perhaps also including either facial recognition and textual cues ~\citep{Bertini2005, Bertini2006} or image segmentation ~\citep{Messelodi2013, Saric2009, YeSPIE2005}. 

Some of the more recent works ~\citep{LuCVPR2011, LuICIMCS2013} attempt to identify players from the field-view shots of sports broadcast videos by modeling appearances with a feature combination of MSER ~\citep{Matas2002}, SIFT ~\citep{Lowe2004}, and color features of players. Deep learning approaches ~\citep{Gerke2015,Gerke2017,Senocak2018} offer promising results, although they have only been used for soccer and basketball player identification on single frames. Most recently, a coarse-to-fine jersey number recognition framework providing discriminative features ~\citep{Zhang2020} is proposed for identifying basketball players and uses a multi-camera setup. Our goal is to build upon these deep learning approaches and identify hockey players from only a single camera broadcast video.

\section{Methodology}

\subsection{Approach}
\begin{figure}[ht]
\centering
\includegraphics[width=0.6\textwidth]{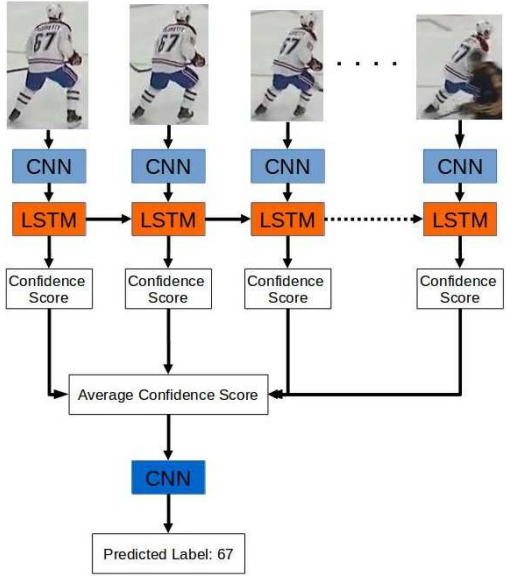}
\caption{Flowchart of our proposed approach. Each frame in the tracklet feeds into the ResNet+LSTM network to generate a confidence score for that frame. The confidence scores are combined by arithmetic mean and fed into a secondary 1D-CNN classifier to output the predicted label for the tracklet.}
\label{dataset_generation_flowchart2}
\end{figure}

To identify hockey players, we propose the ResNet+LSTM network, an end-to-end trainable deep neural network in a supervised learning setting, that combines a residual network (ResNet) ~\citep{He2015} as the CNN base and a recurrent long short-term memory (LSTM) ~\citep{Hochreiter1997} layer on top. We use this network to process an input sequence of cropped bounding box images of a player and output a confidence score for each image. We then identify the player in the input sequence with a late fusion technique. This fusion step combines all the confidence scores from the image sequence via the arithmetic mean and classifies the average confidence score with a secondary CNN classifier to generate one final label prediction for the sequence. Our approach is summarized in a flowchart in Figure~\ref{dataset_generation_flowchart2}.

The nature of videos allows us to determine the identity of hockey players by making observations over multiple instances. With our network, we identify players solely based on the visual features of the jersey number without taking into consideration any priors, such as game or player context. For example, a player’s time-on-ice information is a context that would allow us to know exactly when the player was on the ice at any given moment during a game. This context can help narrow down the list of possible candidates when performing player identification during a game, but these data were not available. 

\begin{figure}[ht]
\centering
\includegraphics[width=0.9\textwidth]{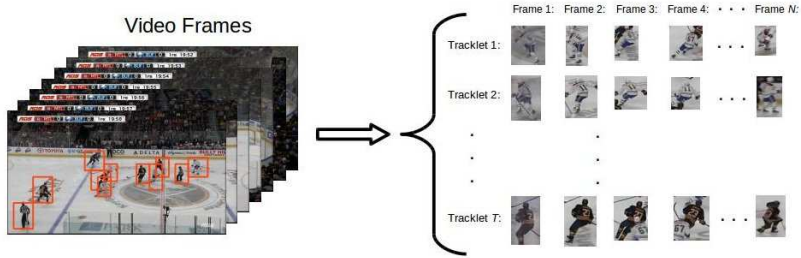}
\caption{Bounding boxes of players are cropped directly from consecutive NHL broadcast hockey video frames to produce player tracklets. The video frames pictured are from the main camera of the rink.}
\label{frames}
\end{figure}

In order for us to observe a player over a period of time, we acquire Sportlogiq’s player tracking data to detect the players in every video frame and mark their pixel location with bounding boxes, as shown in Figure~\ref{frames}. The estimated detections are associated with similar detections across consecutive frames to infer trajectories or tracklets of each player. In this work, we define a \textbf{player tracklet} as a fragment of a player’s overall trajectory and consists of a bounding box image of the player at every time-step of the tracklet. These tracklets can be produced as image sequences of players and can vary in length, since players are constantly appearing and disappearing from the main camera's field of view. For our experiments, we annotate and categorize these tracklets to create a new dataset to train and evaluate the ResNet+LSTM network.

\subsection{Datasets}
To perform player identification, we require bounding box images of hockey players cropped directly from NHL broadcast hockey videos with jersey number annotations. Although there are no publicly available datasets of this sort at the time of this work, we were able to obtain access to the Sportlogiq NHL Hockey Player (NHL-HP) dataset\footnotemark, which contains single images of players. Since our proposed network operate on sequential data, we also acquired Sportlogiq's player tracking data and NHL broadcast hockey videos to generate tracklets to create the NHL Hockey Player Tracklet (NHL-HPT) dataset\footnotemark[\value{footnote}].
\footnotetext{These datasets are only available for research purposes and can be obtained by contacting Sportlogiq.}

\subsubsection{Sportlogiq NHL Hockey Player Dataset}
The Sportlogiq NHL Hockey Player (NHL-HP) dataset is a collection of individual hockey player bounding box images annotated with the player’s jersey numbers that range from 1 to 98 (number 99 has been retired league-wide to honor Wayne Gretzky). These bounding box images are cropped directly from broadcast hockey video frames. The dataset only contains player images when the jersey number on the back of the player is fully visible and identifiable by annotators, without partial views or any parts of the jersey number occluded or hidden. The dataset is divided into two sets, with 45,469 images for training set and 3,885 images for testing. This total also includes augmented data via cropping, padding and scaling.

\subsubsection{NHL Hockey Player Tracklet Dataset}
\begin{figure}[tp]
\centering
\includegraphics[width=\textwidth]{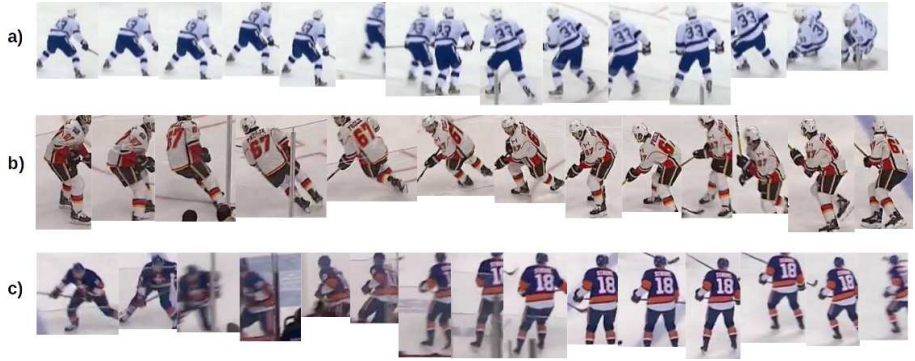}
\caption{Shown are examples of player tracklets with visible jersey numbers. Tracklets are sampled from videos at 30fps. Above, the images displayed are sampled from the tracklets at 3fps.}
\label{visible_frames}
\end{figure}

\begin{figure}[h]
\centering
\includegraphics[width=\textwidth]{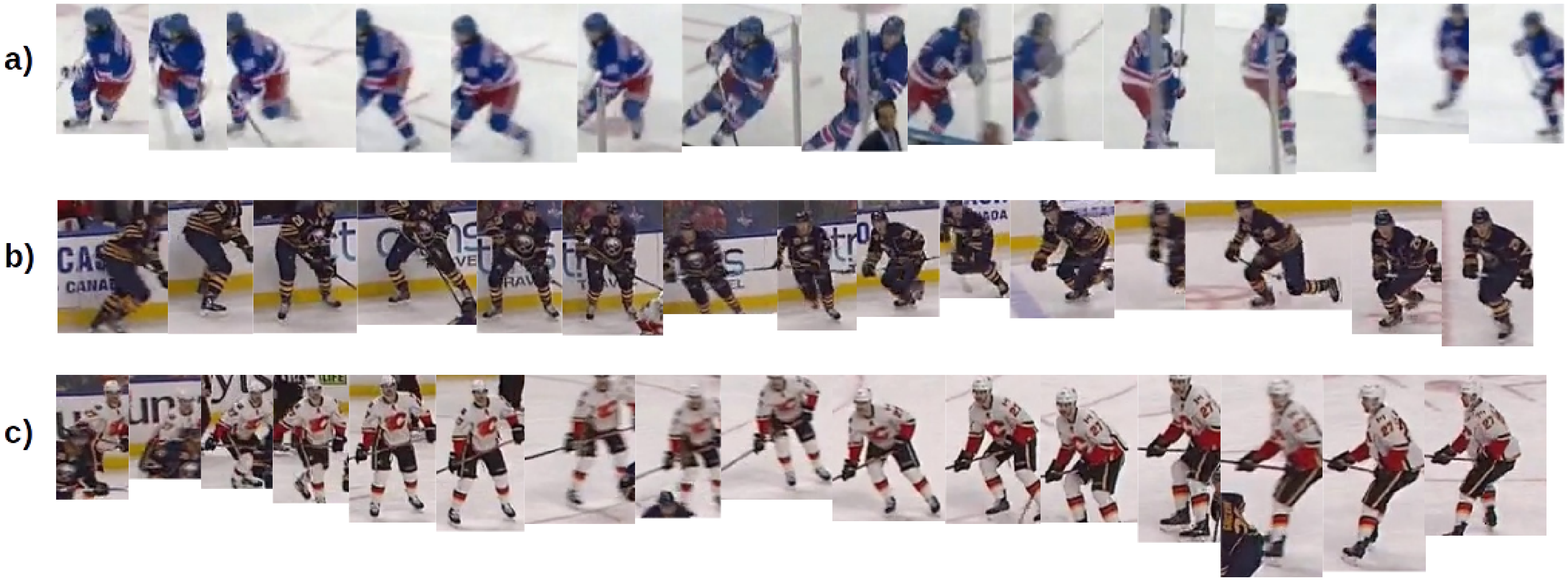}
\caption{Shown are examples of player tracklets with non-visible jersey numbers. Tracklets are sampled from videos at 30fps. Above, the frames displayed are sampled from the tracklets at 3fps.}
\label{nonvisible_frames}
\end{figure}

The dataset currently contains tracklet examples for 81 classes, which includes 79 different jersey numbers, referees, and unidentifiable tracklets. The length of the tracklets in the dataset can vary considerably from 16 to several hundred images. All images in a tracklet are annotated with the same label. The labels for each tracklet are determined by inspection of the back jersey number in all images in the tracklet. Note that a player’s jersey number may not be visible in every image in a tracklet. Figure~\ref{visible_frames} shows examples of three player tracklets that have visible jersey numbers. In case a jersey number is not visible throughout an entire tracklet, we reserve an ‘unknown’ label for these unidentifiable tracklets, as shown in Figure~\ref{nonvisible_frames}. Numbers also exist on the arm sleeve of the jerseys, but these numbers are usually too small to be recognizable and are also considered as ’unknown’. Thus, the models are only trained to predict the back jersey number. Oftentimes, a player’s jersey number is not always visible to viewers, so assigning a label to confirm a player is unknown is more appropriate than classifying the player with an incorrect label.

\begin{figure}[htbp]
\centering
\includegraphics[width=\textwidth]{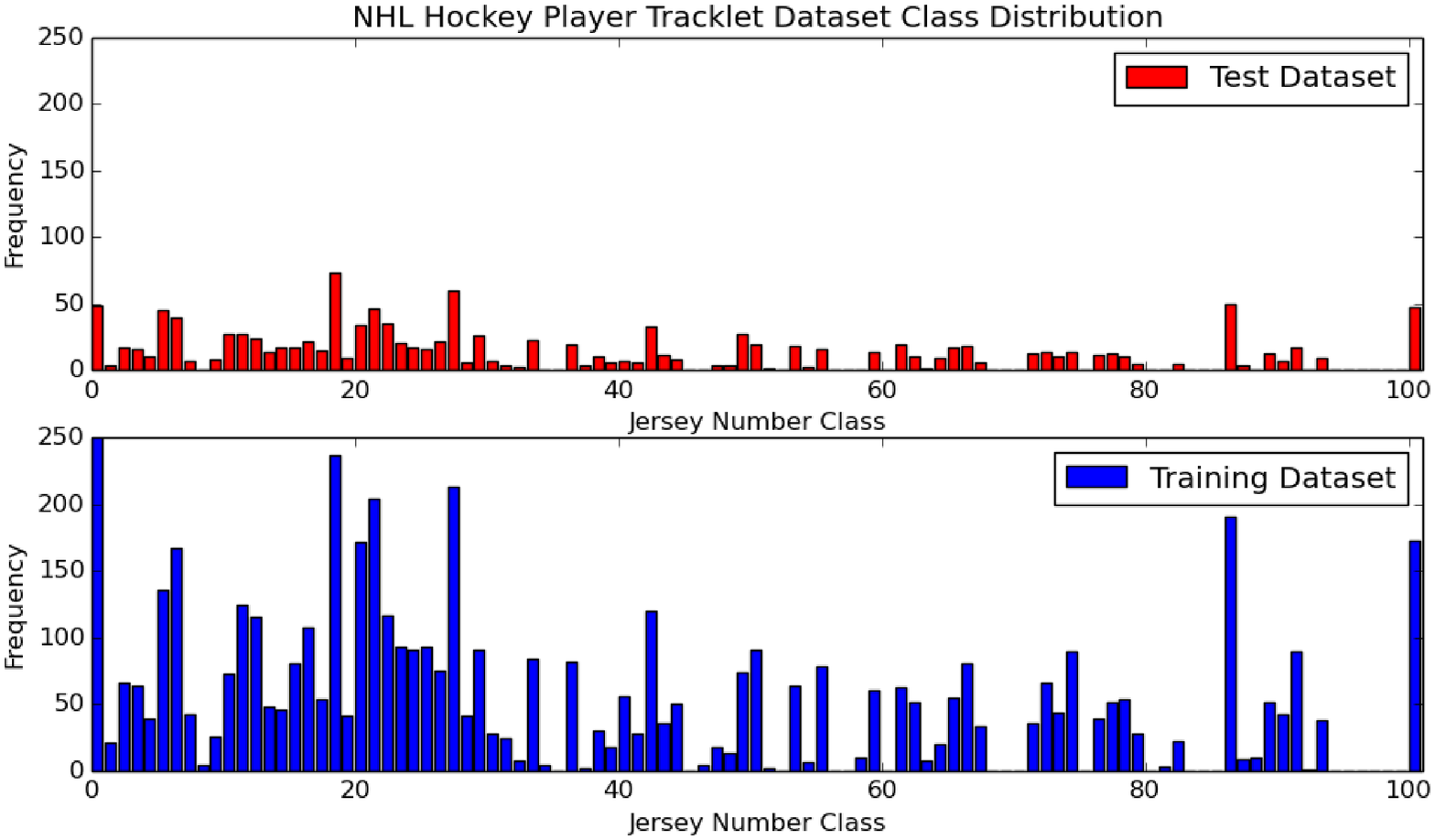}
\caption{The plots show the distribution of number of examples (frequency) in each class in both the test and training subsets of the NHL Hockey Player Tracklet Dataset. The distribution is unbalanced due to the uneven availability of data for some jersey numbers. The label ‘0’ is allocated for referees and the label ‘100’ is for unknown player tracklets where the jersey number is not visible at all. All other labels correspond to the actual jersey number.}
\label{tracklet_data_distribution}
\end{figure}

The dataset currently does not cover all possible jersey numbers in the NHL as there are not enough data available during the creation of the dataset. We randomly divide the dataset into two sets: a training set with 5,267 tracklet examples, and a test set with 1,278 tracklet examples. Due to the uneven availability of data for each jersey number, our dataset is unbalanced, as displayed in the class distribution plot of the training and test dataset in Figure~\ref{tracklet_data_distribution}. This creates a challenge for our network to learn all the visual features that distinguish jersey numbers from each other. 

\subsection{Method}

\subsubsection{ResNet + LSTM Architecture}
We adopt the ResNet-10 \citep{Simon2016} as the CNN base of the network in the architecture of our proposed ResNet+LSTM network. This is due to its relatively shallow architecture for faster training and runtime and high performance compared to the other CNNs we experimented with, which aids in executing experiments. However, note that the full system is designed as an offline solution since the final player number of the player tracklet can only be predicted by the secondary CNN classifier after the fusion of the partial bounding box predictions. 

All layers of the ResNet base, up to the last average pooling layer, are kept the same as the original ResNet-10 \citep{Simon2016}. The input frames first pass through a batch normalization layer, which eliminates the need for mean subtraction during training. The first convolutional layer feeds into a batch normalization layer, a ReLU activation function, and a max pooling layer. After that, there are four residual 2-layer stacks, which were first introduced in \cite{He2015} for non-deep residual networks. Each stack consists of two convolutional layers with a batch normalization layer and a ReLU activation function in between the two convolutional layers. At the end of each stack is an element-wise sum of the second convolutional layer and the input of the stack via a shortcut connection. The residual learning of each stack can be defined as $y = F(x, {W_i})+x$. Here, $x$ and $y$ are the input and outputs of the stack. The function $F(x, {W_i})$ describes the mapping to be learned over a series of layers. For the 2-layer stacks in our network, $F(x, {W_i}) =W_2\sigma(W_1x)$, where $\sigma$ represents the batch normalization and ReLU function and $W_1$ and $W_2$ denotes the weights of the first and second convolutional layers. However, \(y = F(x, {W_i})+x\) is only computable if the dimensions of $x$ and $F$ are equal. In the second, third, and fourth stack, a projection shortcut is required to match the dimensions of $x$ with that of $F$. This is done by performing a linear projection $W_s$ on the input of the stack with 1x1 convolutions. Thus, the residual learning of each stack is defined as $y = W_2 \sigma(W_1x)+W_sx$.

The stack output then feeds into another batch normalization layer and ReLU activation function. The last stack feeds into an average pooling layer, where the pooling is performed globally over each feature map to reduce its spatial size to 1x1. The output of the average pooling layer is a 512-dimensional vector representation of an image, which feeds into a LSTM layer. The LSTM layer has 256 hidden units and receives a series of 16 512-dimensional vectors as an input sequence from the CNN layers. We apply a dropout rate of 0.5 to the LSTM output before feeding into a fully connected layer with a softmax activation function. The dropout technique sets the output of a hidden unit to zero with a probability of 0.5 to limit the number of activations during the backpropagation training process. The softmax classifier outputs an 81-dimensional vector (for 81 classes in our dataset) as a confidence score to predict the label for each input frame. The architecture of our ResNet+LSTM model is summarized in Table~\ref{resnet_lstm}. 

\begin{table}[h]
\centering
\begin{tabular}{ |c|c| } 
\hline
Input & $240\times320\times3$ $\rightarrow$ $227\times227\times3$ \\
\hline
Conv1 & $7\times7$, 64, stride 2, pad 3 \\
\hline
Max Pool & $3\times3$, stride 2 \\
\hline
\multirow{2}{5em}{Conv2\_x} & $3\times3$, 64, stride 1, pad 1\\
& $3\times3$, 64, stride 1, pad 1 \\ 
\hline
\multirow{2}{5em}{Conv3\_x} & $3\times3$, 128, stride 2, pad 1\\ 
& $3\times3$, 128, stride 1, pad 1 \\
\hline
\multirow{2}{5em}{Conv4\_x} & $3\times3$, 256, stride 2, pad 1\\ 
& $3\times3$, 256, stride 1, pad 1 \\ 
\hline
\multirow{2}{5em}{Conv5\_x} & $3\times3$, 512, stride 2, pad 1\\ 
& $3\times3$, 512, stride 1, pad 1 \\ 
\hline
Avg Pool & global pool \\
\hline
LSTM & 256 hidden units, sequence of 16 frames \\
\hline
Dropout & ratio = 0.5 \\
\hline
FC & 81 classes \\
\hline
\end{tabular}
\caption{Network architecture of the ResNet+LSTM model excluding batch normalization and ReLU activation functions. Hyperparameters including filter size, number of filters, and stride are summarized for the convolutional layers. The convolutional layers are grouped by their respective residual layer stacks.}
\label{resnet_lstm}
\end{table}

\subsubsection{Late Fusion Scheme}
To predict a single label for the entire player tracklet $T_s$ with $N$ frames, we consider all $N$ confidence scores predicted for each frame in the tracklet by the ResNet+LSTM network. We use a late score-level fusion method to combine all the confidence scores to output a more accurate label prediction for the tracklet. Let us denote $M$ as the number of classes in the tracklet dataset with $S$ tracklet examples. Each frame label confidence score is a vector $F = [p_1, p_2, ... , p_M]$, where each $p_m$ denotes the confidence score for the $m$th label to be correct. For a tracklet with $N$ frames, the ResNet+LSTM network will output $N$ confidence scores where the score for the $n$th frame is: $F_n = [p_n^1, ... , p_n^m, ... , p_n^M] \epsilon R^M$.


In our late fusion approach, we first calculate the average confidence score for each tracklet by averaging the probabilities for each label across all the frames. Since there are 81 different class labels, the average confidence score is a 81-dimensional vector given as $F_{avg} = [p_1, p_2, ... , p_M] \epsilon R^M$, where $M=81$. Instead of viewing the class with the maximum likelihood in $F_{avg}$ as the predicted label, we can predict the label by providing $F_{avg}$ as a feature vector input to a secondary supervised classifier. 

This secondary classifier is a shallow 1-dimensional convolutional neural network (1D-CNN) with 6 convolutional layers and 3 fully connected layers in total. Some of the convolutional layers are arranged such that there are 2 residual stacks within the network. A shortcut connection is formed between the outputs of the second and third convolutional layers to perform element-wise addition, as well as between the fourth and fifth layers. To avoid down-sampling the already small spatial size of the 81-dimensional input vector, we use filter sizes of 1x3 with a stride of 1 and padding of 1 on each side of the vector representation to maintain the spatial size. The first 2 fully connected layers both have 256 neurons and the last fully connected layer has 81 to serve as the softmax classifier. The architecture of the 1D-CNN is summarized in Table~\ref{1dcnn}.

Given $F_{avg}$ as a feature vector input, the 1D-CNN can detect the patterns within $F_{avg}$ and match them to the patterns that commonly occur in the average confidence scores of the corresponding label. For example, the average confidence score for a tracklet labeled as '45' as ground truth may tend to have higher probabilities for jersey number labels with the digits '4' and '5', such as '45', '43', or '15'. This is usually due to the inaccuracies of the player bounding box classification process or occlusion of one of the digits in the frames. Taking advantage of these patterns in the confidence scores has proved to increase the accuracy results by 4.22\%.

\begin{table}[h]
\centering
\begin{tabular}{ |c|c| } 
\hline
Input & $1\times81$ vector \\ 
\hline
Conv1 & $1\times3$, 20, stride 1, pad 1 \\
\hline
Conv2\_1 & $1\times3$, 50, stride 1, pad 1 \\
\hline
Conv2\_2 & $1\times3$, 50, stride 1, pad 1 \\
\hline
Eltwise Sum & Conv2\_1+Conv2\_2 \\
\hline
Conv3\_1 & $1\times3$, 70, stride 1, pad 1 \\
\hline
Conv3\_2 & $1\times3$, 70, stride 1, pad 1 \\
\hline
Eltwise Sum & Conv3\_1+Conv3\_2 \\
\hline
Conv4 & $1\times2$, 70, stride 1, pad 1 \\
\hline
Max Pool & $1\times2$, stride 2 \\
\hline
FC1 & 256 hidden units \\
\hline
ReLU & $\downarrow$ \\
\hline
FC2 & 256 hidden units\\
\hline
ReLU & $\downarrow$\\
\hline
FC3 & 81 classes\\
\hline
\end{tabular}
\caption{Complete network architecture of the late fusion 1D-CNN classifier}
\label{1dcnn}
\end{table}

\subsection{Training Methodology}
\subsubsection{ResNet+LSTM Training}
All of our model training uses backpropagation to update the network parameters. We use the publicly available Caffe model of ResNet-10 \citep{Simon2016}, pre-trained on the ImageNet \citep{Deng2009} dataset, as the base of our ResNet+LSTM network. The training process first involves pre-training the ResNet base before training the entire network from end-to-end. We train the ResNet-10 base on a subset of the Street View House Number (SVHN) dataset \citep{Netzer2011}. Our subset of the SVHN dataset contains 16,222 examples that are categorized by 10 classes, where each class is associated with a potential digit. By first training on the SVHN dataset, we can improve the weight initialization and generalization of the ResNet-10 for the task of identifying jersey numbers. Once pre-trained, we fine-tune all the layers in the ResNet-10 on the player bounding box images in the Sportlogiq NHL-HP dataset.

When we train the ResNet+LSTM network from end-to-end, we first transfer the weight parameters from the fine-tuned ResNet-10 to initialize the weights of the ResNet layers in the ResNet+LSTM network. The network is trained on inputs of short sequences of 16 consecutive bounding box frames. Similar to the training method used in \cite{Donahue2016} to mimic data augmentation, each training input sequence of 16 consecutive frames is randomly selected from a tracklet $T_s$ with $N$ frames. This input process creates a higher number of possible training inputs for the network. The frames are resized to 320x240 and a random 224x224 crop is taken from each frame. During the end-to-end training, we freeze the ResNet layers to prevent further updates to the initial weight parameters, which we found in our experiments to significantly improve the performance of our network. 

We aim to optimize the ResNet+LSTM network by testing a range of hyperparameters. Relating to the network structure, we find that similar to \cite{Donahue2016}, a LSTM layer with 256 hidden units, an input size of 512 dimensions, and a dropout rate of 0.5 on the output is optimal. In our training, we achieve our best results with a batch size of 16, momentum of 0.9, and a learning rate of 0.005 using stochastic gradient descent with a step down learning rate policy.

\subsubsection{Late Fusion Training}
We use a shallow 1-dimensional CNN (1D-CNN) as the secondary classifier to predict the overall label for a tracklet, given the tracklet’s confidence score as an input. To train the 1D-CNN, we created a dataset of tracklet confidence scores by collecting the outputs from the ResNet+LSTM network given all the training tracklet examples from the NHL-HPT dataset as inputs. More outputs can be generated from the ResNet-LSTM to improve the training of the 1D-CNN, but that would require more tracklet examples.

In our experiments, the network did not noticeably overfit on the training examples, which is likely due to the randomization of the training process. The ResNet+LSTM network’s ability to generalize well with the tracklet training dataset allows us to use the confidence scores from the training examples as appropriate data to train the 1D-CNN. Since the 1D-CNN is fairly shallow, training time took approximately 20 minutes to run 195 epochs.

\section{Experiments}

\subsection{ResNet+LSTM vs. Baselines}

\subsubsection{Implementation Details}
In our experiments, we benchmark several models on the NHL Hockey Player Tracklet (NHL-HPT) dataset. We compare the performance of our ResNet+LSTM network with two baseline models from which our model's architecture is derived: the ResNet-10 ~\citep{Simon2016} network and the LRCN ~\citep{Donahue2016} model. With each model, we compute the confidence score of each frame in a tracklet and average all of the confidence scores in order to calculate the overall score for the tracklet. The jersey number with the highest score is assigned as the predicted jersey number for that particular tracklet. We evaluate the accuracy performance of each model by computing the percentage of test tracklets with the correct predicted jersey number. All experiments are implemented in the Caffe \footnote{The implementation of Caffe we used was updated to support recurrent networks with LSTM units and is available at https://github.com/BVLC/caffe.} framework and executed on a Tesla K40 GPU.

\textbf{ResNet-10.} Since the CNN base of our proposed ResNet+LSTM architecture is the ResNet-10, we compare the results of our ResNet+LSTM network with just the base to demonstrate the improvements gained by adding an LSTM layer. The ResNet-10 model we use in this experiment is trained on single images in the Sportlogiq NHL-HP dataset. 

\textbf{LRCN.} We follow a training methodology similar to ~\cite{Donahue2016} to train the LRCN on our NHL-HPT dataset as we did for our ResNet+LSTM network. We transferred the weights of its base CNN, trained on the NHL-HP dataset, to the LRCN before we started training with input sequences of 16 consecutive frames randomly selected from a tracklet from the NHL-HPT dataset. We trained the LRCN for 100,000 iterations and used the same hyperparameters as ~\cite{Donahue2016}. For each training iteration, we used a batch size of 24 16-frame input sequences. 

\subsubsection{Results and Discussion}
Our ResNet+LSTM network delivered the highest performance in player tracklet identification with an 82.79\% accuracy rate, followed by the LRCN with an accuracy of 80.83\% and the ResNet-10 with an accuracy of 59.08\%. Compared to a typical CNN like the ResNet-10, the networks with LSTM layers saw a significant boost in performance by an increase in accuracy of 22.14\% for the ResNet+LSTM and 21.75\% for the LRCN. The ResNet+LSTM and LRCN achieved similar results, with the ResNet+LSTM achieving a slightly better performance, due to an improved CNN base. The inference speed of the ResNet+LSTM network is 114.29fps.

\begin{figure}[h!]
\centering
\begin{subfigure}{1\textwidth}
\includegraphics[width=\textwidth]{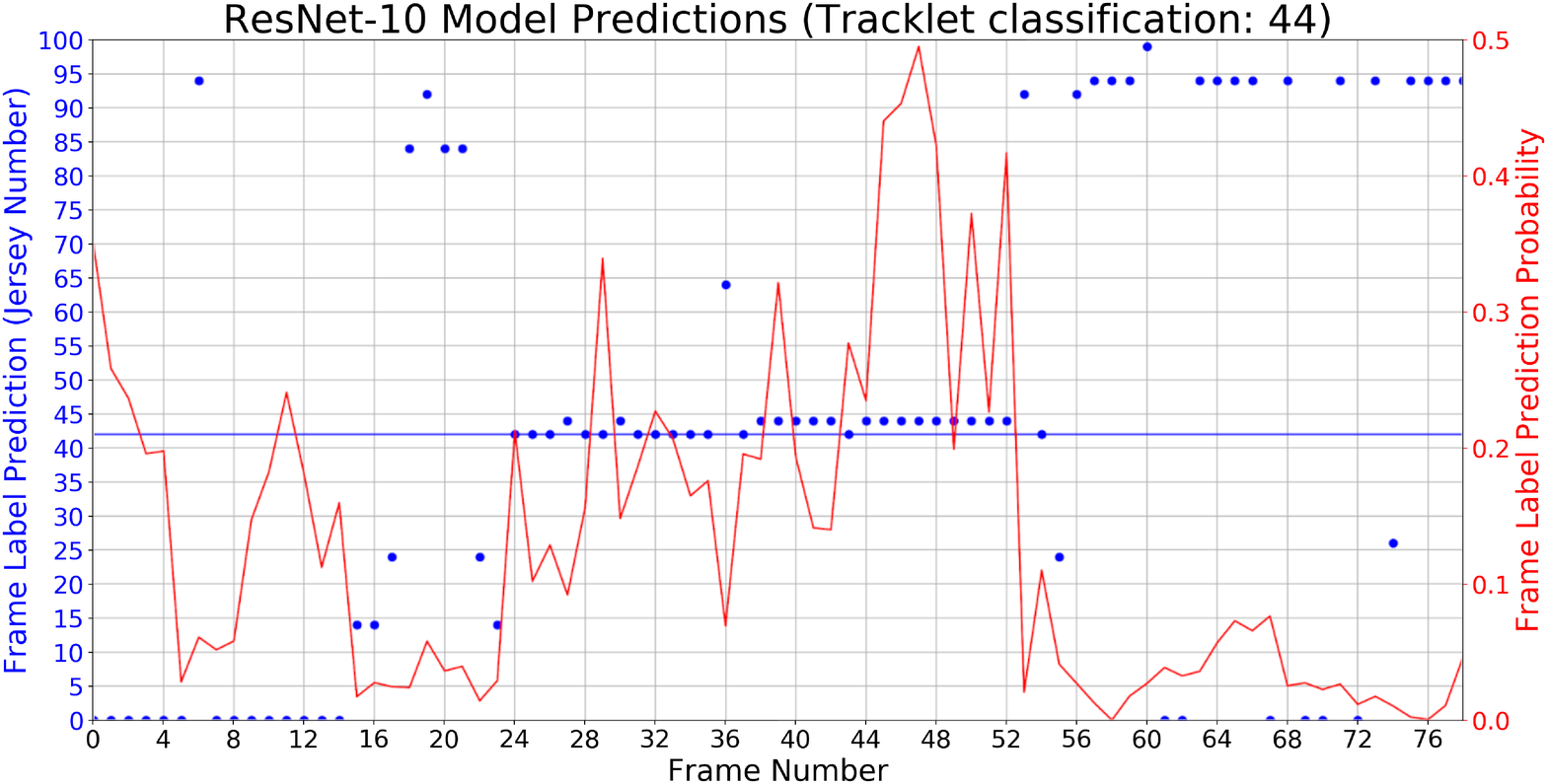}
\end{subfigure}
\begin{subfigure}{1\textwidth}
\includegraphics[width=\textwidth]{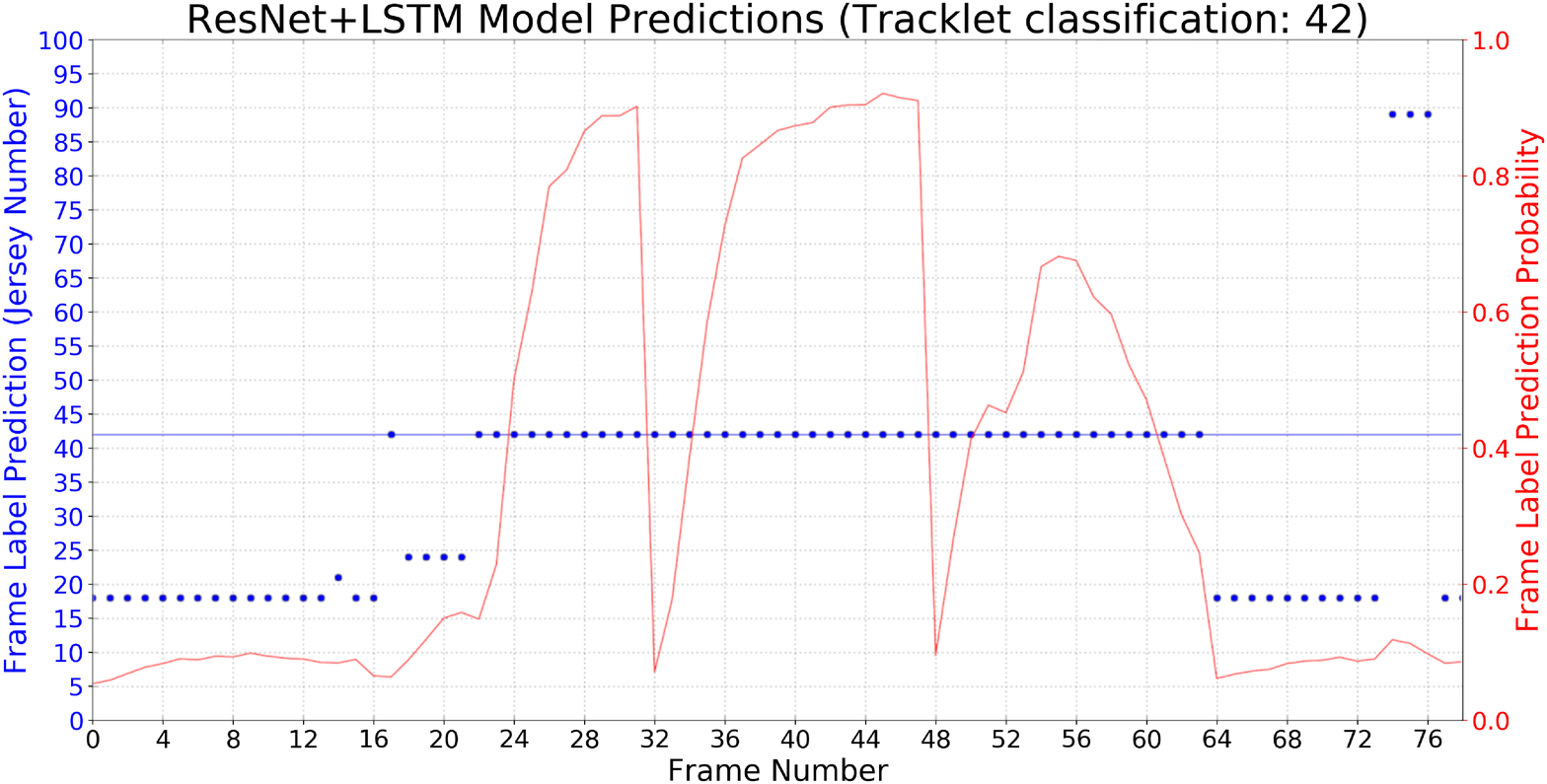}
\end{subfigure}
\caption{Comparison of ResNet-10 (\textbf{top}) and ResNet+LSTM (\textbf{bottom}) label prediction (blue dots) and corresponding confidence probability (red curve) for each frame (along x-axis) in a selected tracklet example. This tracklet is labeled by ground truth as '42' and contains 79 frames. The blue line in both plots marks the ground truth label of the tracklet, which is '42'.}
\label{tracklet_plot}
\end{figure}

For a more qualitative analysis of the difference an additional LSTM layer can make on the performance of a conventional network, we plot the ResNet-10 and ResNet+LSTM networks’ outputs on a frame-by-frame basis to observe their behavior (Figure~\ref{tracklet_plot}). We can see in the ResNet-10 plot (Figure~\ref{tracklet_plot}, top) that the network is able to predict the correct label for several frames in the tracklet. However, the network also incorrectly classified several frames as the label '44' with relatively higher confidence probabilities. Due to the high confidence in the wrongly classified labels, the overall tracklet is incorrectly labeled as '44' instead of the true label, '42'. In the ResNet+LSTM plot (Figure~\ref{tracklet_plot}, bottom), significantly more frames are classified with the correct label than the frames in the ResNet-10 plot. The confidence probabilities of the predicted labels in the ResNet+LSTM plot are also consistently higher for labels that are correctly predicted and lower for labels that are incorrectly predicted. We also noticed in the plot that the confidence probability increases within intervals of 16 frames, such as from the 16th frame to the 31st frame and from the 32nd frame to the 47th frame. Since the LSTM layer operates on input sequences of 16 frames, it can only evaluate sections of 16 frames in a tracklet. Input sequences of greater length may enable better performance since this would allow the LSTM to observe more frames and retrieve more relevant data. However, the constraint on sequence length is placed on the LSTM layer to accommodate the available data because all the tracklets in the NHL-HPT dataset are at least half a second (or 16 frames) or greater in length. At the beginning of each section of 16 frames, the confidence probability is low since the LSTM has not yet observed any previous frames. But as the network evaluates more frames in the section, the LSTM can accumulate information from previous frames in its current hidden state to make a label prediction for the current frame with a higher confidence. This results in the steep confidence probability increase within intervals of 16 frames in the plot. By considering observations from previous frames with the LSTM layer, the ResNet+LSTM is able to accurately predict more correct labels than the ResNet-10 for this particular tracklet example. 

\subsection{Late Fusion Methods}

\subsubsection{Implementation Details}
Up until now, we have been using the arithmetic mean to combine the confidence scores of all the frames in a tracklet to find the overall label for the tracklet. We aim to further improve the performance of our approach by experimenting with different late score-level fusion techniques to observe any changes in performance compared to the arithmetic mean baseline.

To test each late fusion method, we begin by concatenating all $N$ confidence score vectors (of $M$ dimensions) from $N$ frames to produce an $NxM$ confidence score matrix for a given tracklet, as described in \textit{Section 3.4.2}. We experiment with simple mathematical operations across all scores or top $N$ scores to produce a single overall confidence score vector from the matrix. Another method involves treating the average confidence score of a tracklet as a feature vector to be classified by a secondary one-dimensional CNN. 

\subsection{Results and Discussion}
In general, late fusion methods with simple mathematical operations like the median rule, product rule, logarithmic sum, and geometric mean achieved poor results and fared worse than the average rule (82.79\%). Out of all the strategies we attempted, implementing a secondary 1D-CNN classifier to classify a tracklet’s overall confidence score by treating it as a feature vector saw the greatest improvement in accuracy. Initially, we implemented a one-dimensional CNN (1D-CNN v.1) based on the LeNet ~\citep{Lecun1998} architecture and achieved a test accuracy of 85.05\% on the NHL-HPT test set. This boost in performance encouraged us to experiment more with the 1D-CNN approach and we were able to increase the accuracy score of our approach to 87.01\% with the final 1D-CNN (v.2) (See Table~\ref{overall_comparison}). In most CNNs like LeNet, the volume of the data representation is usually down-sampled spatially throughout the network. Since our particular data inputs are already spatially small, down-sampling would cause a loss of data. Our modified 1D-CNN v.2 eliminates max-pooling layers and maintains the spatial size of the input vector with different convolution configurations. The performance increase of the 1D-CNN v.2 can also be attributed to its residual layer structure as opposed to regular convolutional layers.

\begin{figure}[h]
\centering
\includegraphics[width=1\textwidth]{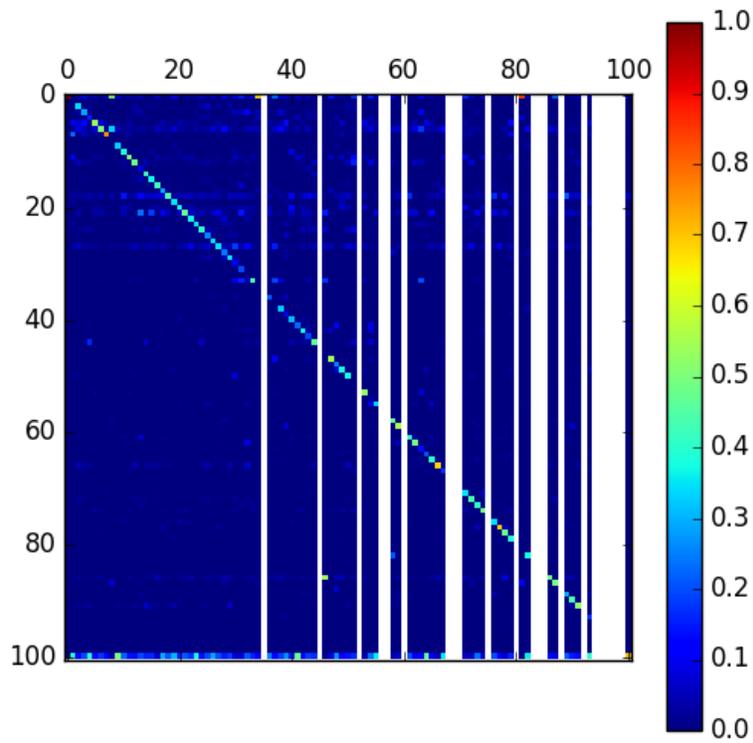}
\caption{Confusion matrix of frame-based classification using ResNet+LSTM. The white "bands" in the matrix represent the jersey numbers that are not available in the NHL-HPT dataset.}
\label{confusion_matrix_frames2}
\end{figure}

\begin{figure}[H]
\centering
\includegraphics[width=1\textwidth]{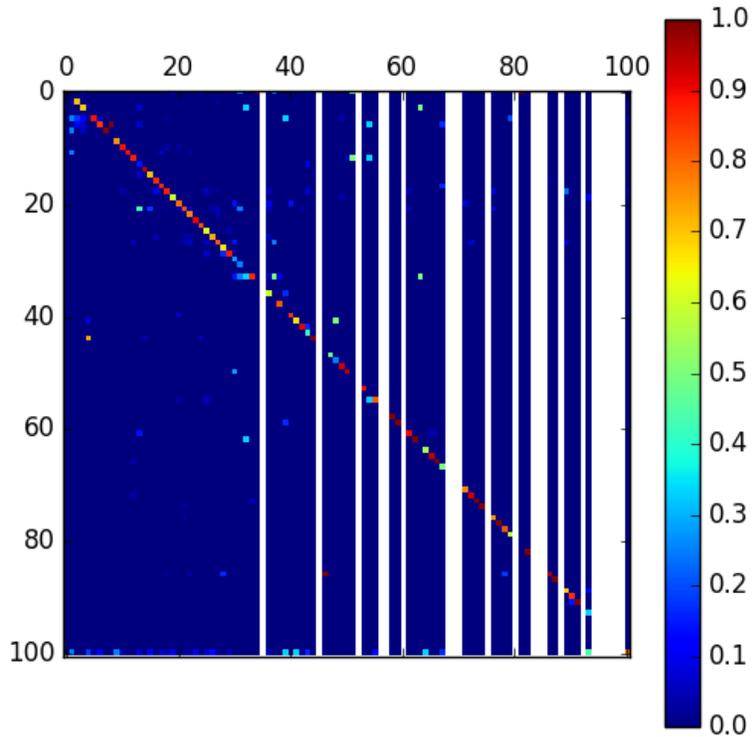}
\caption{Confusion matrix of tracklet-based classification after implementing the 1D-CNN late fusion classifier. The white "bands" in the matrix represent the jersey numbers that are not available in the NHL-HPT dataset.}
\label{confusion_matrix}
\end{figure}

For a closer analysis, we plot a confusion matrix (Figure~\ref{confusion_matrix}) of the tracklet classification results after implementing a secondary 1D-CNN classifier to classify a tracklet’s overall confidence score. We compare this to a confusion matrix (Figure~\ref{confusion_matrix_frames2}) of individual frame prediction results of the ResNet+LSTM network to highlight the advantages of tracklet-based prediction. A clear diagonal can be seen in both plots, which signifies correctly classified entries. The confusion matrix of tracklet classification shows a consistent improvement in accuracy along the diagonal compared to frame-based classification. There are many frame instances labelled as 'unknown' (class label '100') since jersey numbers are not always visible, but when classifying tracklet instances, the model is able to identify the correct jersey number far more often.

However, a small percentage of false classifications can still be seen scattered around the diagonal in Figure~\ref{confusion_matrix}. Some jersey number classes, such as '32', '51', '54', and '63', experience more confusion than others. Upon closer inspection of the dataset class distribution plot in Figure~\ref{tracklet_data_distribution}, these classes have far less numbers of training examples available. Clusters can be seen in the matrix, which suggests that these numbers were predicted as another number that is chronologically adjacent or close. This can be explained since adjacent numbers usually share the same tens digit. This sort of confusion appears to be the main source of error for our system. A more apparent cluster is seen around the single digit numbers area, particularly for classes '1' to '4', where these single digit numbers are incorrectly predicted as other single digit numbers. Visually, it is easier to mislabel a single digit number with another single digit number when players are blurry, so the errors are understandable. Class '4' however, is also mislabelled as double digit labels '40' and '44', but this can be explained since they share the same digits. A few jersey numbers between '30' and '70' are sometimes seen to be predicted as a single digit number corresponding to one of its digits. During the classification process, one of the digits may have been more prevalent than the other throughout the tracklet. Overall, the confusion matrix shows a strong performance in hockey player identification, considering the limited size of our dataset.

\begin{table}[h]
\centering
\begin{tabular}{ |c|c| } 
\hline
\textbf{Models} & \textbf{Test Accuracy} \\
\hline
ResNet-10~\citep{Simon2016} & 59.08\% \\
\hline
LRCN~\citep{Donahue2016} & 80.83\% \\
\hline
ResNet+LSTM (ours) & 82.79\% \\
\hline
ResNet+LSTM \& 1D-CNN v.1 (ours) & 85.05\% \\
\hline
ResNet+LSTM \& 1D-CNN v.2 (ours) & 87.01\% \\
\hline
\end{tabular}
\caption{Overall methods comparison table. Accuracy results are based on their performance on the NHL-HPT test set.}
\label{overall_comparison}
\end{table}

\section{Conclusion}

In this work, we address the task of hockey player identification in NHL broadcast videos and present our recurrent convolutional neural network based approach for jersey number recognition. Compared to previous work in sports player identification, we establish this task as a sequential problem involving multiple frames of a player as inputs. The recurrent nature of our model has shown to boost performance compared to the naive approach of operating on each frame individually. Additionally, we employ a late score-level fusion method of using a secondary 1-dimensional CNN classifier to classify the overall tracklet confidence score to achieve a higher accuracy score of 87.01\%.

\section*{References}

\bibliography{reference}

\end{document}